\documentclass[11pt, a4paper]{googledeepmind}

\pdftrailerid{redacted}

\makeatletter
\renewcommand\bibentry[1]{\nocite{#1}{\frenchspacing\@nameuse{BR@r@#1\@extra@b@citeb}}}
\makeatother

\usepackage{dsfont}

\usepackage[authoryear, sort&compress, round]{natbib}

\usepackage{defs}

\newboolean{shownotes}
\setboolean{shownotes}{false}

\title{Open Problems and a Hypothetical Path Forward in LLM Knowledge Paradigms}

\author[*]{

\Authfont{
Xiaotian Ye$^{1}$,
Mengqi Zhang$^{2}$,
Shu Wu$^{3}$
}

\Affilfont{
$^1$School of Computer Science, Beijing University of Posts and Telecommunications, $^2$Shandong University \\
$^3$New Laboratory of Pattern Recognition (NLPR), State Key Laboratory of Multimodal Artificial Intelligence Systems (MAIS), Institute of Automation, Chinese Academy of Sciences
}

\texttt{yexiaotian@bupt.edu.cn, mengqi.zhang@sdu.edu.cn, shu.wu@nlpr.ia.ac.cn} 
}

\begin{abstract}

Knowledge is fundamental to the overall capabilities of Large Language Models (LLMs). The knowledge paradigm of a model, which dictates how it encodes and utilizes knowledge, significantly affects its performance. Despite the continuous development of LLMs under existing knowledge paradigms, issues within these frameworks continue to constrain model potential.

This blog post highlight three critical open problems limiting model capabilities: (1) challenges in knowledge updating for LLMs, (2) the failure of reverse knowledge generalization (the reversal curse), and (3) conflicts in internal knowledge.
We review recent progress made in addressing these issues and discuss potential general solutions. 
Based on observations in these areas, we propose a hypothetical paradigm based on Contextual Knowledge Scaling, and further outline implementation pathways that remain feasible within contemporary techniques. Evidence suggests this approach holds potential to address current shortcomings, serving as our vision for future model paradigms. 

This blog post aims to provide researchers with a brief overview of progress in LLM knowledge systems, while provide inspiration for the development of next-generation model architectures.
\end{abstract}

\begin{document}
\captionsetup[subfigure]{justification=centering}

\maketitle

\section{Introduction}
\label{sec:introduction}

Large language models (LLMs) pre-trained on massive datasets have achieved widespread success in various downstream tasks and real-world applications \citep{openai2023gpt}. These pre-trained LLMs demonstrate the ability to make predictions based on real-world facts and, with further fine-tuning, can incorporate this knowledge to perform reasoning across complex problems and diverse tasks \citep{Bubeck2023-gm}. In essence, LLMs can be viewed as models that encode and leverage knowledge. 

The knowledge embedded in these models is crucial to their performance on downstream tasks. Solving complex problems, particularly those involving reasoning, fundamentally depends on prior knowledge; models cannot solve complex tasks from scratch without any prior knowledge. This highlights the importance of the knowledge paradigm -- that is, the way in which models acquire, store, and utilize knowledge.  
% While the current paradigm has achieved remarkable progress through scaling \citep{openai2023gpt}, it faces inherent limitations. 
While the capabilities of models under current paradigm have been continuously improved through scaling \citep{openai2023gpt}, such improvements are inevitably limited by inherent issues of the paradigm itself.
As high-quality real-world training data becomes increasingly scarce and the pace of performance gains slows \citep{villalobos2024position}, identifying and addressing these limitations becomes essential to unlocking the full potential of LLMs.

In this blog post, we begin with a brief overview of the existing knowledge paradigm within LLMs (\S \ref{sec:preliminaries}). We then summarize three open problems that limit model capabilities, arising from the current LLM knowledge paradigm, and review recent progress in addressing them (\S \ref{sec:problems}). Subsequently, we discuss several potential general solutions that are emerging across these areas (\S \ref{sec:general_solution}). Building on these insights, we speculate on future model paradigms and introduce a proposed hypothetical model design, discussing its possibilities implementation within the current framework (\S \ref{sec:hypothetical_model}).

\section{How do LLMs Model Knowledge: Background and Preliminaries}
\label{sec:preliminaries}

One remarkable capability of LLMs is their ability to store and utilize vast amounts of information, including real-world knowledge. Sequence model architectures, not limited to transformers, but also including RWKV \citep{peng-etal-2023-rwkv} and Mamba \citep{gu2024mamba}, have demonstrated similar levels of knowledge capability after pre-training. This suggests that knowledge acquisition is more closely tied to the training paradigm than to the specific architecture. In this section, we primarily discuss how language models represent knowledge from the perspective of architecture-agnostic training paradigm, providing background and preliminaries for subsequent discussions.

\paragraph{Training Paradigm}
The foundation of LLM knowledge lies in probabilistic language modeling, which constitutes the core loss during the unsupervised pre-training phase of the model \citep{Radford2018ImprovingLU,radford2019language}. For a sample $\mathbf{x}$ containing several tokens $(x_1, x_2, ..., x_n)$, the language model learns to predict the target token $x_i$ based on all preceding tokens $x_{<i}$ by modeling the probability of a specific token in context. In the most prevalent causal language models, the following language modeling objective is the most commonly used, and the training objective is to maximize the following probability:

\begin{equation}
\mathcal{L}_{L M}(\mathbf{x})=\sum_{i=1}^{n} \log P\left(x_{i} \mid \mathbf{x}_{<i}\right)
\end{equation}

Pre-training constitutes the primary knowledge acquisition phase \citep{DBLP:journals/corr/abs-2404-05405}. During the pre-training phase, LLMs are trained on a massive corpus. This training objective enables the model to perform generative tasks by autoregressively predicting the next token. Numerous experiments have shown that large models memorize a significant amount of knowledge during this stage, which can also be considered the primary phase for knowledge acquisition \citep{akyurek2022towards,cao-etal-2024-retentive-forgetful}.

\paragraph{Knowledge Modeling Paradigm}

Through this probabilistic approach, LLMs naturally learn information in the form of conditional probabilities, such as $p\left(s_{n-k}, \ldots, s_{n} \mid s_{1}, \ldots, s_{n-k-1}\right)$. This process allows the model to capture statistical relationships between tokens, including real-world factual information present in the corpus. For example, when trained on a sentence like ``\textit{The president of the United States in 2024 is Biden}'', the model learns to predict ``Biden'' based on the preceding context, increasing the probability $p \left(\texttt{Biden} \mid \texttt{The president...is} \right)$. \citet{radford2019language} noted that this training paradigm facilitates transfer to downstream tasks because tasks can be framed within the same probabilistic framework, e.g., $p (output \mid input)$. For instance, a factual question-answering task can be expressed as $p (answer \mid question)$, leveraging the patterns learned during pre-training.

Learning language modeling on fact-related corpora is essentially the process by which the model acquires knowledge, this aligns with the definition of knowledge in LLMs. Knowledge is an awareness of facts, and a fact can be expressed in various forms, such as question-answer pairs or sentences. For LLMs, knowledge is most commonly defined as the model possessing knowledge k if it can correctly answer the corresponding question $r_k$ \citep{wang-etal-2024-knowledge-mechanisms}. In simpler terms, an LLM's knowledge is its capability to produce factual outputs for inputs related to the real world. This is a performance-based definition, focusing on the observable behavior of the model: as long as it demonstrates factual input-output mappings or $p (input \mid output)$, it is considered to possess knowledge. And this is exactly what language modeling accomplishes.

Within this framework, knowledge in LLMs is represented in a highly implicit manner. The model’s knowledge capabilities are encoded and manifested entirely through language modeling. Abstracting away from architectural specifics, all knowledge is captured and leveraged as statistical associations between tokens, which are learned and stored within black-box parameters. We first informally summarize the current paradigm of how LLMs model knowledge as follows:

\begin{tcolorbox}[title=Current LLM Knowledge Paradigm (informal), colback=orangecolor!5!white,colframe=orangecolor!85!black]
LLMs encode factual information into model weights through pre-training based on probabilistic language modeling, forming input-output mappings that statistically capture real-world relationships in the training data.
\end{tcolorbox}

Such paradigm can be described as a ``Probabilistic Language Modeling-based Paradigm.'' Many practical issues related to LLM knowledge stem from this modeling approach, which we will analyze in subsequent sections.

\section{Open Problems and Advancements}
\label{sec:problems}

The current knowledge paradigm of LLMs introduces several unique challenges. Given the crucial role of an LLM's knowledge base in various downstream tasks, addressing these issues is vital for further advancements in LLM capabilities. In this section, we will discuss several open problems.

\subsection{Challenges in Updating LLM Knowledge}
\label{subsec:knowledge-edit}

Many real-world applications of LLMs, especially knowledge-intensive tasks, require the models to possess accurate and up-to-date information. This creates a practical need for efficiently and cost-effectively updating the knowledge embedded within LLMs \citep{zhang2024comprehensivestudyknowledgeediting}. However, it is not that straightforward under current knowledge paradigms.

Simply put, when we want to update the knowledge in an LLM, we hope to change the LLM's behavior to satisfy the following three criteria \citep{zhang2024comprehensivestudyknowledgeediting,yao-etal-2023-editing}: (i) accurately respond to new knowledge, (ii) preserve existing knowledge without catastrophic forgetting, and can (iii) leverage updated knowledge in complex reasoning tasks. The biggest difficulty in this task is closely related to the paradigm by which models construct knowledge. 
Under the current paradigm, the knowledge exhibited by LLMs emerges from the probabilistic modeling of correlations between tokens. This modeling approach intuitively lacks interpretability, and the knowledge is entirely encoded within the model's weights. Consequently, modifying this knowledge requires adjusting the model's black-box parameters -- a challenging process that inevitably encounters issues such as catastrophic forgetting.

Knowledge editing \citep{zhang2024comprehensivestudyknowledgeediting} has been proposed as a strategy for knowledge updating, yet the effectiveness of existing methods remains unsatisfactory. 
Knowledge editing methods have largely been successful in addressing first two requirements, but they encounter significant obstacles in addressing the third, \textbf{\textit{reasoning ability}}, which has become the core focus of knowledge editing work in the past two years \citep{zhang2025uncovering,zhong-etal-2023-mquake}. For example, if we want to edit a counterfactual fact into the LLM, such as ``\textit{Microsoft was founded by Steve Jobs},'' we also want the model to automatically apply this fact to the reasoning of related questions, such as ``\textit{Which college did the founder of Microsoft attend?}'' \citep{,cohen-etal-2024-evaluating} This involves multi-hop reasoning that includes the new fact. The generalization performance on this task is far from satisfactory: for instance, using the representative multi-hop reasoning dataset MQuAKE \citep{zhong-etal-2023-mquake}, the SOTA editing method that achieves 99.7\% accuracy when editing the counterfactual knowledge itself in Vicuna-7B, but only achieves 6.9\% accuracy when asked multi-hop reasoning questions. An important observation in this field is that \textit{it is extremely difficult to adjust parameters to enable the model to acquire sufficiently generalized knowledge from very few samples.}

Furthermore, it remains questionable whether efficient knowledge updating is truly achievable within the current knowledge paradigm. The inner workings of LLMs are themselves a controversial topic, and some unconfirmed issues about these mechanisms are fatal to the task of knowledge updating: one such question is whether LLMs genuinely perform latent reasoning \citep{biran-etal-2024-hopping,yang-etal-2024-large-language-models}. To illustrate, consider the earlier question about the university attended by the founder of Microsoft. Does the model first extract the fact that Microsoft's founder from its parameters and then connect this to the second hop to infer the university, or does it simply memorize the entire multi-hop question as a separate fact? If the latter is true, knowledge updating can never be achieved in a simple way, since changes to base facts cannot propagate. Although it is difficult to obtain clear results through black-box interpretability analysis, this possibility cannot be ruled out: some studies suggest that evidence of such reasoning is unstable and  heavily influenced by fact composition types and context \citep{yang-etal-2024-large-language-models}. These indicate that modifying parameters for knowledge updating remains a challenging problem with a long road ahead.

\subsection{The Reversal Curse in Knowledge Generalization}

Another significant issue with the current LLM knowledge paradigm is the so-called ``Reversal Curse''. This phenomenon reveals that the current method of encoding knowledge prevents models from generalizing to very simple logical extensions of the training samples.

% 介绍问题定义
The ``Reversal Curse'' phenomenon was initially identified by \citet{berglund2024the}: If a model is trained on a sentence of the form ``A is B'', it will not automatically generalize to the reverse direction ``B is A''. For example, if the training corpus contains ``A is B's parent,'' the model cannot automatically answer ``B is A's child.'' It's important to note, however, that if the information ``A is B'' is present within the LLMs' context, the model can infer the reverse relationship. \citet{berglund2024the} demonstrated in their original work that this phenomenon is widespread in autoregressive language models and difficult to mitigate -- this is undoubtedly significant, because the model's ability to handle such simple logical generalizations is an important prerequisite for further performance improvements. This phenomenon has also been observed and analyzed in subsequent works\citep{lin2024delving,zhu2025towards}.

The Reversal Curse is often criticized as a limitation of the autoregressive paradigm, representing an inherent flaw within the current knowledge paradigm. Causal models like GPT derive their knowledge from language modeling, which learns \textit{\textbf{conditional probabilities}} from the training corpus. By increasing the probability of output ``B'' given input ``A is,'' the model effectively establishes a probability mapping of $p \left( \texttt{B} \mid \texttt{A is} \right)$. Conditional probability, by its very nature, is unidirectional, and the model architecture itself is designed to learn this unidirectional dependency: the gradients during training on ``A is B'' will change A's representation to include information about B, but will not affect B's representation \citep{berglund2024the}; this makes it difficult for the model to recall complete information about A based solely on B's representation.

A series of subsequent works have emerged to alleviate the reversal curse problem in LLMs, including using bidirectional model editing to attempt to mitigate the issue \citep{ma2024untyingreversalcursebidirectional}, or training the model with the data reversed, excluding entity names \citep{golovneva2024reverse}.  The underlying idea is similar: using data augmentation-like techniques to re-inject the reverse version of each piece of knowledge. While these are partially effective, they still significantly increase costs and fail to fundamentally address the issue, as the root of the problem likely stems from the model architecture itself. Some challengers to the autoregressive paradigm, such as diffusion language models \citep{nie2025largelanguagediffusionmodels}, do not exhibit these problems and highlight this as one of their major advantages.

\subsection{Internal Knowledge Conflicts in LLMs}

Another critical issue stemming from the knowledge paradigm is the presence of knowledge conflicts. Here, to specifically analyze issues related to knowledge paradigm, we will focus on \textbf{internal} knowledge conflicts within the model.

For a pre-trained model, one of the primary source of internal knowledge contradictions arises from biases inherent in the pre-training corpus itself \citep{xu-etal-2024-knowledge-conflicts}. Many current LLM pre-training corpora are derived from web scraping, such as Common Crawl \footnote{\url{https://www.commoncrawl.org/}}. Such corpora contain a significant amount of inherently conflicting information: some information is simply \textit{\textbf{incorrect}} due to the variable quality of internet content, ranging from reliable sources to misinformation and rumors \citep{bian2023influenceexternalinformationlarge}. Furthermore, even \textit{\textbf{factually correct}} information can present conflicting perspectives. We briefly outline three possible scenarios:

\begin{itemize}
    \item For \textbf{time-sensitive information}, answers may vary depending on the time of capture. For example, the answer to ``Who is the Prime Minister of the UK?'' would have been Boris Johnson based on data crawled in 2020, but Sunak based on data from 2022.
    \item Much information is \textbf{subjective and potentially highly controversial}. For example, personal choices like whether or not to donate blood can elicit a wide range of answers when scraped from the web, some of which could be potentially harmful. \citep{weidinger2021ethicalsocialrisksharm}
    \item There are also inherently difficult problems, such as scientific questions, for which there is \textbf{currently no definitive answer or summary statement}, meaning the training corpus may only contain a disorganized collection of viewpoints.
\end{itemize}

How do we expect the model to handle such knowledge? For an ideal case, we might hope it could discern potential falsehoods based on information sources, recognize temporal information and remember what is new, answer subjective questions from a neutral and correct perspective, and, for uncertain complex problems, respond with ``I don't know'' while summarizing the diverse viewpoints as much as possible \citep{wang2024resolving}. In simpler terms, we desire the model to possess the ability to \textit{\textbf{select}} and \textit{\textbf{synthesize}} knowledge during acquisition \citep{xu-etal-2024-knowledge-conflicts}. These capabilities are natural to humans and are reflected in our knowledge acquisition process.

However, considering that current models are based on \textbf{probabilistic modeling}, this ideal scenario is not feasible. During pre-training, the model's learning rule is to increase the joint probability of sentences in the corpus -- this process doesn't differentiate the content of the corpus, and typically lacks awareness of data sources or temporal information. Considering knowledge in the form of one-to-many question-answer pairs, when multiple answers $(A_1,A_2, ... , A_n)$ exist for the same question $Q$, the pre-training process encourage the model to increases all $P(A_i \mid Q)$, rather than selecting and summarizing the answers as we ideally expect. 

This characteristic suggests that \textit{probabilistic modeling may not be ideally suited for directly handling one-to-many knowledge that involves inherent conflicts or requires reasoning and selection}, unless knowledge is pre-processed and structured before being modeled probabilistically. Otherwise, the trained model might learn to output either \textit{several answers with similar probabilities at random}, or, the \textit{most frequently occurring answer in the pre-training corpus} \citep{lu-etal-2024-scaling}. In practical scenarios, this can cause the model to provide uncertain answers and allow errors and harmful information from the pre-training corpus to contaminate the generated results \citep{xu-etal-2024-knowledge-conflicts, chang-bergen-2024-language}. This poses a significant challenge to both the performance and safety of the model.

Current research offers some solutions for specific cases mentioned above, such as using safety alignment to prevent the model from outputting harmful content \citep{ji2023ai} or cleaning the training data \citep{grattafiori2024llama3herdmodels}. However, some issues caused by probabilistic modeling still remain unsolved. At a minimum, the model is still unable to summarize contradictory knowledge in the training data as we would like, and the safety alignment is not robust enough: these strategies can be bypassed by methods such as jailbreaking \citep{yi-etal-2024-vulnerability}. In addition, it is worth noting that research on context-related knowledge conflicts has proven that LLMs have the ability to identify contradictory knowledge in the context, and summarizing different viewpoints in the input is also a common ability of LLMs \citep{xu-etal-2024-knowledge-conflicts}. Knowledge conflicts under context may be easier to deal with than internal knowledge conflicts.

\section{Is a General Solution Possible?}
\label{sec:general_solution}

The previous section summarized the issues with existing pre-trained knowledge-based paradigms and subsequent research advancements. In this section, we will summarize several common threads among these solutions, as potential, albeit imperfect, general solutions.

\subsection{Synthesic Data Generation}

Many issues encountered in LLMs can be traced back to the training data, thus a natural idea is to incorporate sufficient data augmentation and data cleaning into the training data \citep{allenzhu2024physicslanguagemodels31}. This represents a potentially universal solution to the aforementioned problems.

\paragraph{Potential}
Data augmentation through synthetic data generation is a common post-training strategy. It is already a widely applied and practically important technique for improving a model's specific capabilities \citep{grattafiori2024llama3herdmodels}. However, the situation with the aforementioned problems is somewhat different. Typically, data augmentation doesn't require processing every possible sample; a certain number of augmented samples are sufficient for the model to generalize to similar tasks. In contrast, the problems discussed above are closely related to the fundamental knowledge paradigm of LLMs, and augmenting only a subset of the knowledge may still leave other knowledge areas constrained. For example, the reversal curse has been observed that it will not be alleviated by such augmentation \citep{berglund2024the}.

Despite these challenges, the data augmentation strategy still has the potential to address these issues. If we consider an extremely ideal condition, where we don't need to consider any implementation costs, data augmentation alone would be sufficient: (1) The biggest obstacles to \textbf{\textit{updating knowledge}} are generalization and the retention of irrelevant knowledge. If we augment a large amount of relevant data, and even re-train with irrelevant knowledge on the scale of pre-training, we can certainly satisfy these two points. (2) For the \textbf{\textit{reversal curse}}, simply adding the reverse versions of facts to the corpus \citep{golovneva2024reverse}, doubling the data size, should be feasible. And for (3) \textbf{\textit{knowledge conflict}}, we can summarize the conflicting knowledge in the corpus in advance using other models, retaining only the standard answer we want for training the model.

\paragraph{Limitations}
However, in practice, one of the biggest factor to consider is exactly the implementation cost. Latest LLMs have remarkably large pre-training corpora; for example, Llama 3 \citep{grattafiori2024llama3herdmodels} was pre-trained on 15T tokens. Cleaning and augmenting such a vast corpus is a huge undertaking, and it is likely that the benefits would not be proportional to the effort: we have only considered three issues of the existing paradigm above, and each issue already requires a separate data augmentation strategy. The reversal curse alone would necessitate doubling the number of training tokens -- and thus the cost. This doesn't even account for the myriad other potential problems, \textit{each} of which would introduce further overhead if addressed through data augmentation.

Furthermore, given that the problems we have outlined stem from the knowledge paradigm itself, data augmentation alone offers only a partial solution and does not address the root cause. This makes the increased cost seem even less justifiable. In summary, while synthetic data offers certain advantages, its potential remains constrained by many inherent limitations.

\subsection{In-Context Learning}
\label{subsec:in-context-learning-generalsolution}

Another interesting observation from the progress in these areas is that these issues are largely absent when using in-context learning. This makes it another potential universal solution.

\paragraph{Potential}
In-context learning has been considered a promising direction in  \textbf{\textit{knowledge editing}}, largely due to its few-shot generalization capabilities. When new knowledge is provided as context to the LLM, it demonstrates a remarkable ability to leverage this contextual information across a variety of tasks, and to utilize it in highly complex reasoning \citep{cohen-etal-2024-evaluating}. This has inspired one type of knowledge editing method, in-context editing, which chooses to change the model output through external modules and often utilizes in-context learning \citep{zhang2024comprehensivestudyknowledgeediting}: many methods are applied in a way that is more like enhancing the effect of Retrieval-Augmented Generation (RAG) \citep{karpukhin-etal-2020-dense}. This approach, which avoids manipulating complex black-box parameters, circumvents many problems and, in fact, far surpasses all parameter modification methods in reasoning tasks\citep{zhang2024comprehensivestudyknowledgeediting,zhong-etal-2023-mquake,cohen-etal-2024-evaluating}. Overall, a key observation in is that the few-shot generalization capability of the in-context mechanism is considerably stronger than parameter modification, thereby enabling efficient and generalizable knowledge updates.

For the other two issues, in-context learning can also serve as a solution. Regarding the \textbf{\textit{reversal curse}}, \citet{berglund2024the} noted in the original paper that identified this phenomenon that there were no such generalization problems for information provided in the context: the reversal curse occurs because unidirectional conditional probability learning tasks make it difficult for models to retrieve information related to preceding tokens based on subsequent tokens in the training corpus. However, contextual knowledge is provided completely to the model without involving such process, thus avoiding this issue. Concerning \textbf{\textit{knowledge conflicts}}, existing research shows that LLMs are capable of discerning whether viewpoints within the context are contradictory, and can distill and summarize them \citep{xu-etal-2024-knowledge-conflicts} -- this is in fact one of the most common use cases for LLMs in document analysis. These observations align with our everyday experiences using large language models.

\paragraph{Limitations}
Despite its potential, in-context learning is often not considered a serious solution in practice. Although contextual knowledge avoids these issues, such prompting techniques are generally difficult to deploy as part of a model in practice, often requiring additional modules like RAG to become usable. Nevertheless, RAG introduces external overhead, and the performance ceiling is strongly limited by the retriever's capabilities \citep{yao-etal-2023-editing}. Furthermore, for more complex problems such as asking an LLM to freely write based on a given proposition, the knowledge involved can be vast, and it is thus difficult to ensure the retriever retrieves all potentially relevant and up-to-date knowledge. In summary, despite the promising mechanisms of in-context learning, the key question is how can we better leverage it.

\section{Hypothetical Model with Contextual Knowledge Scaling}
\label{sec:hypothetical_model}

The previous section discussed the open problems, research progress, and common threads among existing solutions within the current LLM knowledge paradigm. In this section, we will conjecture potential solutions building on these insight. We begin by presenting several inferences and conjectures derived from these observations (\S\ref{subsec:method-background}), then introduce our hypothetical model design (\S\ref{subsec:method}) and outline potential implementation approaches (\S\ref{subsec:method-imple}).

\subsection{What Can We Learn from These Areas?}
\label{subsec:method-background}

\paragraph{Observation}
We previously discussed three open problems in the LLM knowledge paradigm in Section \ref{sec:problems}: knowledge updating, the reversal curse, and knowledge conflicts. Section \ref{sec:general_solution} explored two potential general solutions. Combining these challenges and corresponding approaches, we can summarize the following key observations, which are also the points we want to highlight:

\begin{itemize}
\item Knowledge is encoded as an input-output mapping within the model weights. When the number of training sample is insufficient, it becomes difficult to learn sufficiently generalized knowledge through parameter updates alone, limiting the model’s ability to handle complex tasks. This also complicates efficient knowledge updating under the current paradigm.

\item Knowledge is modeled implicitly through probabilistic association. This implicitness gives rise to several intractable issues in recording knowledge, such as the reversal curse and knowledge conflicts, which may hinder generalization and ultimately constrain model capabilities.

\item As a potential solution, in-context learning mechanism exhibits a remarkable generalization ability; pre-trained models are capable of generalize from few-shots by leveraging contextual information, which makes effective knowledge updating possible. It also possesses enhanced robustness: many problems inherent in current knowledge paradigm are significantly less pronounced, or almost non-existent, in in-context learning.
\end{itemize}

\paragraph{Inferences}
Here, we aim to summarize the phenomena discussed above into more conclusive statements, while also incorporating the findings previously mentioned to make some inferences. This may not be entirely rigorous, but it can at least be considered a well-founded conjecture:

\begin{itemize}
\item Given the same amount of data, \textbf{in-context learning is likely to exhibit superior generalization capabilities} compared to approaches that encode the same data into model parameters via language modeling loss used in pre-training.
\item Encoding and utilizing information through \textbf{probabilistic language modeling mechanisms present more robustness challenges,} such as the reversal curse and knowledge conflicts, compared to mechanisms in in-context learning.
\end{itemize}

In simpler terms, in-context learning mechanisms in LLMs seem to possess many advantages in recording and utilizing knowledge, compared to directly encoding it into model weights via probabilistic modeling. Of course this is not to say that it can replace the pre-training and language modeling; they remain indispensable, as many other capabilities of LLMs are built upon them, including instruction following, reasoning, and in-context learning itself. These capabilities clearly require encoding through input-output mapping, making it difficult to construct only input context to enable the model to possess them.

However, this suggests that \textbf{there exists some capabilities that cannot be perfectly encoded by the probabilistic language modeling paradigm}, even though the model exhibits some of these capabilities. For example, for the factual knowledge we are discussing, the probability-based method is at least not very good at handling conflicting knowledge; and in-context learning seems to be a better mechanism at this time. We summarize our core conjecture as follows:

\begin{tcolorbox}[title=Hypothesis 1, colback=citecolor!5!white,colframe=citecolor!75!black]
Certain types of knowledge or capabilities, such as those involving conflicting factual information, are difficult to encode or generalize effectively using the probabilistic language modeling paradigm. For these capabilities, the mechanism of in-context learning offers the potential for improved generalization and may achieve greater data efficiency.
\end{tcolorbox}

This is framed as a hypothesis because it is difficult to confirm its validity in all scenarios. But many practices can mutually verify this, including the problems we discussed earlier, and the common use of RAG to eliminate hallucinations and get better answers. We continue to make the following inferences based on this conjecture.

\paragraph{Further Conjecture}
Assuming the previous conjecture holds true, naturally, we would consider how to best leverage the advantages of this mechanism to further enhance the capabilities of LLMs. The current primary way to utilize the in-context mechanism is to use RAG to retrieve information and then explicitly append it to the context. However, as mentioned earlier, this is limited by the capabilities of the retriever and is inflexible. Here, to fully exploit this capability, we pose the following questions:

\begin{tcolorbox}[title=Research Question, colback=orangecolor!5!white,colframe=orangecolor!85!black]
Given that pre-trained large language models exhibit highly efficient mechanisms for utilizing context information, can we enable this ability to \textbf{access and utilize the entirety of their acquired knowledge}, effectively scaling up to the full pre-training corpus? If so, would such model demonstrate more robust and generalizable knowledge capabilities, potentially surpassing the limitations of conventional paradigms?
\end{tcolorbox}

We will first illustrate this concept and its advantages with a hypothetical model (\S \ref{subsec:method}), and then discuss in more depth what this means in the current model architecture and how to implement it (\S \ref{subsec:method-imple}).

\subsection{Hypothetical Model with Contextual Knowledge Scaling}
\label{subsec:method}

In this subsection, we will initially disregard implementation feasibility, and discuss what such a model would be like and what advantages it would offer. The most direct way to allow a model to access information via the in-context mechanism is to \textit{directly add it all to the context as input}, we will thus use such a hypothetical ``Corpus-in-Context'' model for illustration here:

Suppose we have a pre-trained LLM with an \textbf{infinite context length} and \textbf{unlimited in-context learning capabilities}. Further, assume that during inference, it \textbf{prioritizes contextual knowledge over parametric knowledge}. We use this as our base model and thus we are able to directly provide the entire pre-training corpus of the LLM, or data of equivalent scale, explicitly as its input context.

The context of such a model can be seen as containing all the knowledge required for downstream tasks. While the model weights, like those of a standard LLM, can be considered to encode knowledge, the model prioritizes adherence to contextual knowledge. This means that factual knowledge stored in the parameters can be considered unused in practice, allowing for the simplification that all of the model's knowledge exists in contextual form.
The LLM component is responsible for performing in-context learning, retrieving and processing information within the context to complete downstream tasks. This work may more akin to complex pattern matching and reasoning on the input based on natural language -- interestingly, there is also a current view that LLMs are just performing pattern matching. This might be more explicit in such a model, as all knowledge is explicitly presented in the input for it to operate.

To better understand the implications and differences compared to current approaches, let us formalize this hypothetical paradigm.

\newcommand{\knowledgebase}{\mathcal{K}}
\newcommand{\basemodel}{\mathcal{F}}

\subsubsection{Formalizing the Paradigm}
 The core idea is to leverage the in-context mechanism to its theoretical limit by providing \textit{all} necessary information directly as input context. Assume we use the pre-training corpus $\mathcal{D}_{\text{pretrain}}$, or data of equivalent scale, represented as the context $\knowledgebase$. Such that $\knowledgebase \sim \mathcal{D}_{\text{pretrain}}$.

\paragraph{Pre-training}
In this setup, we assume a base model $\basemodel$ with parameters $\boldsymbol{\theta}_{\text{base}}$. These parameters can be obtained through pre-training on a large corpus $\mathcal{D}_{\text{pretrain}}$ by minimizing a loss function $\mathcal{L}$, similar to standard LLM pre-training involving next token prediction. Specifically,
\begin{equation}
\boldsymbol{\theta}_{\text{base}} \approx \arg\min_{\boldsymbol{\theta}} \sum_{\mathbf{x} \in \mathcal{D}_{\text{pretrain}}} \mathcal{L}(\mathbf{x}; \boldsymbol{\theta})
\end{equation}
\begin{equation}
\mathcal{L}(\mathbf{x}; \boldsymbol{\theta}) = - \sum_{i=1}^{|\mathbf{x}|} \log P(x_i \mid x_{<i}; \boldsymbol{\theta}).
\end{equation}
However, the primary focus here is on the capability gained through pre-training to process and reason over long contexts.

\paragraph{Inference (Knowledge Utilization)}
At inference time, given an input query $x$, the model $\basemodel$ processes the full context  $\knowledgebase \oplus x$. The model then generates an output $y$ as:
\begin{equation}
y \sim \basemodel_{\boldsymbol{\theta}_{\text{base}}}(\knowledgebase \oplus x) \quad \text{s.t.}  \knowledgebase \succ_{\text{know}} \boldsymbol{\theta}_{\text{base}},
\end{equation}
where  $\knowledgebase \succ_{\text{know}} \boldsymbol{\theta}_{\text{base}}$ indicates that the model prioritizes contextual knowledge $\knowledgebase$ over the parametric knowledge encoded in 
$\boldsymbol{\theta}_{\text{base}}$. This prioritization is critical for handling contextual-parametric knowledge conflicts, ensuring that the model responds in accordance with the explicit context. Conceptually, this allows for a simplified conceptualization where the model is viewed as if all its knowledge is stored within the context $\knowledgebase$.

\paragraph{Context Maintenance and Knowledge Update}

The model's full context $\knowledgebase \oplus x$ contains the knowledge base $\knowledgebase$, serving as the source of knowledge, and the more dynamic query $x$. Context updates, typically seen in standard LLMs, here correspond to providing a new query $x$ alongside the existing knowledge base $\knowledgebase$. When the model's knowledge needs updating, new knowledge $\knowledgebase_{\text{new}}$ can be directly incorporated into the knowledge context $\knowledgebase$, i.e.,
\begin{equation}
\knowledgebase \leftarrow \knowledgebase \oplus \knowledgebase_{\text{new}}.
\end{equation}
Conversely, the standard LLM paradigm requires pre-training-like parameter updates on $\boldsymbol{\theta}_{\text{base}}$ for knowledge updating.

\subsubsection{Advantages}

In this hypothetical model, since most knowledge is represented explicitly in the context, the problems caused by knowledge modeling in current paradigm that we discussed earlier no longer exist. This leads to several clear advantages:

\begin{itemize}
\item \textbf{Efficient Knowledge Updating}: New knowledge can be incorporated simply by appending it to the existing large-scale knowledge context. Since all knowledge is externalized in this way, updates become straightforward and benefit from the generalization capabilities of in-context learning.

\item \textbf{Elimination of the Reversal Curse}: The reversal curse is inherently caused by the unidirectional relationships between tokens established by language modeling. In-context learning avoids this problem because the model has access to the complete information (\S\ref{subsec:in-context-learning-generalsolution}). 
\item \textbf{Conflict-aware Knowledge Integration}: Contextual knowledge can be enriched with metadata such as timestamps and source information. The LLM also has the ability to discern and summarize conflicting knowledge within the context, avoiding the problem of conflicting knowledge overwriting each other during pre-training.
\end{itemize}

Furthermore, if the efficiency of utilizing data as context remains higher than that of injecting knowledge into the model weights via language modeling loss, then this paradigm is likely to have a higher data utilization rate than the traditional paradigm. This has the potential to alleviate the current problem of limited high-quality data by improving the utilization of existing data.

So far, we have discussed a hypothetical model that achieves several potential advantages by scaling up the context. We now turn to the question of whether such a model is feasible.

\subsection{How Can We Implement It?}
\label{subsec:method-imple}

The hypothetical model described above does not consider the practical aspects of model architecture. If we consider the mainstream transformer LLMs, it is almost impossible to implement, and the biggest obstacle is the context length: the computational complexity of the mainstream transformer architecture is quadratic in the sequence length, making the cost scale up rapidly as the context grows. This forces us to focus on the model architecture itself. Here, we first analyze the hypothetical model from the perspective of the sequence model mechanism, and then introduce the most promising current ideas.

\paragraph{Contextual Knowledge as Hidden State}

Reconsidering our previous conjecture, our goal is to enable the model to efficiently access a large amount of the information, especially factual knowledge, using a mechanism similar to in-context learning. An intuitive approach is to model the knowledge directly as context. However, the context in sequence models doesn't necessarily require explicit tokens; it is parameterized internally and can be \textbf{compressed and approximated}. As long as the model can access this information using an in-context-learning-like mechanism, we can leverage the advantages of this approach.

Such compression mechanisms are indeed quite natural in sequence modeling. In fact, for more general sequence models like RNNs and beyond, parameterizing and compressing input information is inherently part of their operational design.
As sequence models process input, they record it into a hidden state; this implies that \textbf{\textit{contextual knowledge is equivalent to the hidden state}} in such models. More specifically, the context mechanism of all sequence models can be regarded as storing context tokens $(x_1, x_2, ... , x_t)$ into their hidden state $s_t$. For example, RNNs compress the context into a fixed-size state, and the KV cache of transformers can act as a hidden state similar to RNN, but without any compression \citep{sun2024learninglearntesttime}.

From this perspective, model pre-training can be seen as learning how to update the hidden state and utilize it for prediction. The inference is the process of continuously updating and utilizing $s_t$ using the mechanism learned during training.
Considering the equivalence of hidden states and contextual tokens, \textit{\textbf{the advantage of in-context learning we discussed earlier is equivalent to the advantage of the pre-trained model's mechanism for updating and utilizing hidden state information.}} Interestingly, the learning process of such hidden state mechanism in models like transformers does not craft from human prior; it naturally emerges through self-supervised learning in the process of next token prediction \citep{sun2024learninglearntesttime}. This might be one of the reason for the efficiency of this mechanism.

\textbf{Therefore for sequence models, to achieve our goal, it can be conceptualized as pre-filling their hidden state with an extremely large amount of information in advance. This can be seen as an alternative form of pre-training on hidden state.} While typically regarded as transient buffer for user input, the hidden state itself is actually a place where information can be stored, and our previous analysis shows that its information utilization mechanism may be more efficient. To leverage the advantages of the context mechanism, it is possible to conceptualize pre-filled hidden states as part of the model's weights, and to integrate the process of pre-filling knowledge into the hidden state within the standard model training pipeline. More concisely, we present it as another hypothesis:

\begin{tcolorbox}[title=Hypothesis 2, colback=citecolor!5!white,colframe=citecolor!75!black]
The hidden state of a sequence model can be pre-filled as a potentially more efficient module for storing and accessing information. Pre-filled hidden states can be considered as part of the model's parameters and are directly leveraged during inference.
\end{tcolorbox}

This may not applicable to models like transformers that do not perform hidden state compression, but is more suitable for models with fixed-size hidden states; otherwise it will be difficult to record so much information and distribute it like a pre-trained model. If we can have a good enough model architecture, which has a large enough hidden state and a mechanism for updating and compressing information, then it is expectable that we can obtain a good enough approximation of our hypothetical model through pretraining of the hidden state. Fortunately, recent advances in sequence models offer a promising path towards it.

\paragraph{Efficient Architectures for Long-Context Modeling}

What kind of architecture can meet the above requirements? First, the hidden state needs to have a mechanism for compressing information; otherwise, the state size and computational complexity will increase rapidly with the context length like transformers. Second, it needs to be expressive enough, or have a large enough hidden state and an efficient compression mechanism; otherwise, it will quickly encounter an information bottleneck like traditional RNNs. Many recent modern linear recurrent models can serve as potential solutions, such as Mamba \citep{gu2024mamba}, RWKV \citep{peng-etal-2023-rwkv}, and further, TTT \citep{sun2024learninglearntesttime} and Titans \citep{behrouz2024titanslearningmemorizetest}, whose hidden states are more expressive.

The latest generation of sequence models, represented by TTT and Titans, are likely closer to our expectations. Their key idea is to make the hidden state itself a machine learning model, and the update rule is a step of self-supervised learning: for example, the original TTT paper proposes two implementations, respectively using a linear model and a two-layer MLP \citep{sun2024learninglearntesttime}, which can achieve stronger representation and compression capabilities than ordinary RNN states. These early attempts at new sequence architectures are far smaller in training scale and number of parameters than the powerful pre-trained transformers in the industry, but have already demonstrated promising long-context capabilities. 

A major advantage of making the hidden state a model itself is that the complexity of the hidden state can be high enough to accommodate a sufficient amount of knowledge information. Another, and perhaps more important point is that it can naturally scale up by increasing both its parameter count and architectural complexity, which can be used as potential scaling perspectives in the future. While current model architectures may not be the optimal solutions, future models and hybrid architectures alike hold the potential to realize our hypothesized model, or to achieve comparable or even superior capabilities through alternative approaches. We present the preceding analysis as a projection of potential future model development, and we hope that future models will unlock greater levels of intelligence.

\section{Conclusion}

This blog post provides an overview of the current knowledge paradigm in language models, discussing how they model knowledge and the open problems arising from the limitations of this approach. We summarize existing solutions and several general strategies, and theoretically discuss a hypothetical model with contextual knowledge scaling as a potential avenue for future model development. 

To summarize the viewpoints discussed in this blog post, we propose two conjectures regarding potential improvements to the LLM knowledge paradigm:  (1) In-context learning demonstrates certain advantages over traditional LLM knowledge modeling paradigm, and could potentially be scaled up to the pre-training corpus level to enable models to acquire stronger and more robust knowledge capabilities; (2) The hidden states of sequence models may offer a highly generalizable mechanism for knowledge encoding and utilizing, and could potentially serve as a major knowledge storage module. This module could be pre-filled with a large amount of knowledge as an integral part of the model, and serve as another potential scaling direction.

We framework our analysis as a hypothesis, and hope this blog post will enable readers to grasp the discoveries and progress in the field of LLM knowledge, while offering insights for future model development.

% \section*{Acknowledgements}

\bibliographystyle{abbrvnat}
\bibliography{main}

\begin{thebibliography}{37}
\providecommand{\natexlab}[1]{#1}
\providecommand{\url}[1]{\texttt{#1}}
\expandafter\ifx\csname urlstyle\endcsname\relax
  \providecommand{\doi}[1]{doi: #1}\else
  \providecommand{\doi}{doi: \begingroup \urlstyle{rm}\Url}\fi

\bibitem[Aky{\"u}rek et~al.(2022)Aky{\"u}rek, Bolukbasi, Liu, Xiong, Tenney, Andreas, and Guu]{akyurek2022towards}
E.~Aky{\"u}rek, T.~Bolukbasi, F.~Liu, B.~Xiong, I.~Tenney, J.~Andreas, and K.~Guu.
\newblock Towards tracing factual knowledge in language models back to the training data.
\newblock \emph{arXiv preprint arXiv:2205.11482}, 2022.

\bibitem[Allen{-}Zhu and Li(2024)]{DBLP:journals/corr/abs-2404-05405}
Z.~Allen{-}Zhu and Y.~Li.
\newblock Physics of language models: Part 3.3, knowledge capacity scaling laws.
\newblock \emph{CoRR}, abs/2404.05405, 2024.
\newblock \doi{10.48550/ARXIV.2404.05405}.
\newblock URL \url{https://doi.org/10.48550/arXiv.2404.05405}.

\bibitem[Allen-Zhu and Li(2024)]{allenzhu2024physicslanguagemodels31}
Z.~Allen-Zhu and Y.~Li.
\newblock Physics of language models: Part 3.1, knowledge storage and extraction, 2024.
\newblock URL \url{https://arxiv.org/abs/2309.14316}.

\bibitem[Behrouz et~al.(2024)Behrouz, Zhong, and Mirrokni]{behrouz2024titanslearningmemorizetest}
A.~Behrouz, P.~Zhong, and V.~Mirrokni.
\newblock Titans: Learning to memorize at test time, 2024.
\newblock URL \url{https://arxiv.org/abs/2501.00663}.

\bibitem[Berglund et~al.(2024)Berglund, Tong, Kaufmann, Balesni, Stickland, Korbak, and Evans]{berglund2024the}
L.~Berglund, M.~Tong, M.~Kaufmann, M.~Balesni, A.~C. Stickland, T.~Korbak, and O.~Evans.
\newblock The reversal curse: {LLM}s trained on {\textquotedblleft}a is b{\textquotedblright} fail to learn {\textquotedblleft}b is a{\textquotedblright}.
\newblock In \emph{The Twelfth International Conference on Learning Representations}, 2024.
\newblock URL \url{https://openreview.net/forum?id=GPKTIktA0k}.

\bibitem[Bian et~al.(2023)Bian, Lin, Liu, Lu, Zhang, He, Han, and Sun]{bian2023influenceexternalinformationlarge}
N.~Bian, H.~Lin, P.~Liu, Y.~Lu, C.~Zhang, B.~He, X.~Han, and L.~Sun.
\newblock Influence of external information on large language models mirrors social cognitive patterns, 2023.
\newblock URL \url{https://arxiv.org/abs/2305.04812}.

\bibitem[Biran et~al.(2024)Biran, Gottesman, Yang, Geva, and Globerson]{biran-etal-2024-hopping}
E.~Biran, D.~Gottesman, S.~Yang, M.~Geva, and A.~Globerson.
\newblock Hopping too late: Exploring the limitations of large language models on multi-hop queries.
\newblock In Y.~Al-Onaizan, M.~Bansal, and Y.-N. Chen, editors, \emph{Proceedings of the 2024 Conference on Empirical Methods in Natural Language Processing}, pages 14113--14130, Miami, Florida, USA, Nov. 2024. Association for Computational Linguistics.
\newblock \doi{10.18653/v1/2024.emnlp-main.781}.
\newblock URL \url{https://aclanthology.org/2024.emnlp-main.781/}.

\bibitem[Bubeck et~al.(2023)Bubeck, Chandrasekaran, Eldan, Gehrke, Horvitz, Kamar, Lee, Lee, Li, Lundberg, Nori, Palangi, Ribeiro, and Zhang]{Bubeck2023-gm}
S.~Bubeck, V.~Chandrasekaran, R.~Eldan, J.~Gehrke, E.~Horvitz, E.~Kamar, P.~Lee, Y.~T. Lee, Y.~Li, S.~Lundberg, H.~Nori, H.~Palangi, M.~T. Ribeiro, and Y.~Zhang.
\newblock Sparks of artificial general intelligence: Early experiments with {GPT-4}.
\newblock \emph{arXiv}, Mar. 2023.

\bibitem[Cao et~al.(2024)Cao, Tang, Lin, Jiang, Dong, Han, Chen, Wang, and Sun]{cao-etal-2024-retentive-forgetful}
B.~Cao, Q.~Tang, H.~Lin, S.~Jiang, B.~Dong, X.~Han, J.~Chen, T.~Wang, and L.~Sun.
\newblock Retentive or forgetful? diving into the knowledge memorizing mechanism of language models.
\newblock In N.~Calzolari, M.-Y. Kan, V.~Hoste, A.~Lenci, S.~Sakti, and N.~Xue, editors, \emph{Proceedings of the 2024 Joint International Conference on Computational Linguistics, Language Resources and Evaluation (LREC-COLING 2024)}, pages 14016--14036, Torino, Italia, May 2024. ELRA and ICCL.
\newblock URL \url{https://aclanthology.org/2024.lrec-main.1222}.

\bibitem[Chang and Bergen(2024)]{chang-bergen-2024-language}
T.~A. Chang and B.~K. Bergen.
\newblock Language model behavior: A comprehensive survey.
\newblock \emph{Computational Linguistics}, 50\penalty0 (1):\penalty0 293--350, Mar. 2024.
\newblock \doi{10.1162/coli_a_00492}.
\newblock URL \url{https://aclanthology.org/2024.cl-1.9/}.

\bibitem[Cohen et~al.(2024)Cohen, Biran, Yoran, Globerson, and Geva]{cohen-etal-2024-evaluating}
R.~Cohen, E.~Biran, O.~Yoran, A.~Globerson, and M.~Geva.
\newblock Evaluating the ripple effects of knowledge editing in language models.
\newblock \emph{Transactions of the Association for Computational Linguistics}, 12:\penalty0 283--298, 2024.
\newblock \doi{10.1162/tacl_a_00644}.
\newblock URL \url{https://aclanthology.org/2024.tacl-1.16/}.

\bibitem[Golovneva et~al.(2024)Golovneva, Allen-Zhu, Weston, and Sukhbaatar]{golovneva2024reverse}
O.~Golovneva, Z.~Allen-Zhu, J.~E. Weston, and S.~Sukhbaatar.
\newblock Reverse training to nurse the reversal curse.
\newblock In \emph{First Conference on Language Modeling}, 2024.
\newblock URL \url{https://openreview.net/forum?id=HDkNbfLQgu}.

\bibitem[Grattafiori et~al.(2024)Grattafiori, Dubey, Jauhri, Pandey, Kadian, Al-Dahle, Letman, et~al.]{grattafiori2024llama3herdmodels}
A.~Grattafiori, A.~Dubey, A.~Jauhri, A.~Pandey, A.~Kadian, A.~Al-Dahle, A.~Letman, et~al.
\newblock The llama 3 herd of models, 2024.
\newblock URL \url{https://arxiv.org/abs/2407.21783}.

\bibitem[Gu and Dao(2024)]{gu2024mamba}
A.~Gu and T.~Dao.
\newblock Mamba: Linear-time sequence modeling with selective state spaces.
\newblock In \emph{First Conference on Language Modeling}, 2024.
\newblock URL \url{https://openreview.net/forum?id=tEYskw1VY2}.

\bibitem[Ji et~al.(2023)Ji, Qiu, Chen, Zhang, Lou, Wang, Duan, He, Zhou, Zhang, et~al.]{ji2023ai}
J.~Ji, T.~Qiu, B.~Chen, B.~Zhang, H.~Lou, K.~Wang, Y.~Duan, Z.~He, J.~Zhou, Z.~Zhang, et~al.
\newblock Ai alignment: A comprehensive survey.
\newblock \emph{arXiv preprint arXiv:2310.19852}, 2023.

\bibitem[Karpukhin et~al.(2020)Karpukhin, Oguz, Min, Lewis, Wu, Edunov, Chen, and Yih]{karpukhin-etal-2020-dense}
V.~Karpukhin, B.~Oguz, S.~Min, P.~Lewis, L.~Wu, S.~Edunov, D.~Chen, and W.-t. Yih.
\newblock Dense passage retrieval for open-domain question answering.
\newblock In B.~Webber, T.~Cohn, Y.~He, and Y.~Liu, editors, \emph{Proceedings of the 2020 Conference on Empirical Methods in Natural Language Processing (EMNLP)}, pages 6769--6781, Online, Nov. 2020. Association for Computational Linguistics.
\newblock \doi{10.18653/v1/2020.emnlp-main.550}.
\newblock URL \url{https://aclanthology.org/2020.emnlp-main.550/}.

\bibitem[Lin et~al.(2024)Lin, Fu, Liu, Xie, Lin, Wang, Cai, Wu, and Ye]{lin2024delving}
Z.~Lin, Z.~Fu, K.~Liu, L.~Xie, B.~Lin, W.~Wang, D.~Cai, Y.~Wu, and J.~Ye.
\newblock Delving into the reversal curse: How far can large language models generalize?
\newblock In \emph{The Thirty-eighth Annual Conference on Neural Information Processing Systems}, 2024.
\newblock URL \url{https://openreview.net/forum?id=1wxFznQWhp}.

\bibitem[Lu et~al.(2024)Lu, Li, Cheng, Ding, Huang, and Qiu]{lu-etal-2024-scaling}
X.~Lu, X.~Li, Q.~Cheng, K.~Ding, X.~Huang, and X.~Qiu.
\newblock Scaling laws for fact memorization of large language models.
\newblock In Y.~Al-Onaizan, M.~Bansal, and Y.-N. Chen, editors, \emph{Findings of the Association for Computational Linguistics: EMNLP 2024}, pages 11263--11282, Miami, Florida, USA, Nov. 2024. Association for Computational Linguistics.
\newblock \doi{10.18653/v1/2024.findings-emnlp.658}.
\newblock URL \url{https://aclanthology.org/2024.findings-emnlp.658/}.

\bibitem[Ma et~al.(2024)Ma, Gu, Ling, Liu, and Liu]{ma2024untyingreversalcursebidirectional}
J.-Y. Ma, J.-C. Gu, Z.-H. Ling, Q.~Liu, and C.~Liu.
\newblock Untying the reversal curse via bidirectional language model editing, 2024.
\newblock URL \url{https://arxiv.org/abs/2310.10322}.

\bibitem[Nie et~al.(2025)Nie, Zhu, You, Zhang, Ou, Hu, Zhou, Lin, Wen, and Li]{nie2025largelanguagediffusionmodels}
S.~Nie, F.~Zhu, Z.~You, X.~Zhang, J.~Ou, J.~Hu, J.~Zhou, Y.~Lin, J.-R. Wen, and C.~Li.
\newblock Large language diffusion models, 2025.
\newblock URL \url{https://arxiv.org/abs/2502.09992}.

\bibitem[OpenAI(2023)]{openai2023gpt}
R.~OpenAI.
\newblock Gpt-4 technical report.
\newblock \emph{arXiv}, pages 2303--08774, 2023.

\bibitem[Peng et~al.(2023)Peng, Alcaide, Anthony, Albalak, Arcadinho, Biderman, Cao, Cheng, Chung, Derczynski, Du, Grella, Gv, He, Hou, Kazienko, Kocon, Kong, Koptyra, Lau, Lin, Mantri, Mom, Saito, Song, Tang, Wind, Wo{\'z}niak, Zhang, Zhou, Zhu, and Zhu]{peng-etal-2023-rwkv}
B.~Peng, E.~Alcaide, Q.~Anthony, A.~Albalak, S.~Arcadinho, S.~Biderman, H.~Cao, X.~Cheng, M.~Chung, L.~Derczynski, X.~Du, M.~Grella, K.~Gv, X.~He, H.~Hou, P.~Kazienko, J.~Kocon, J.~Kong, B.~Koptyra, H.~Lau, J.~Lin, K.~S.~I. Mantri, F.~Mom, A.~Saito, G.~Song, X.~Tang, J.~Wind, S.~Wo{\'z}niak, Z.~Zhang, Q.~Zhou, J.~Zhu, and R.-J. Zhu.
\newblock {RWKV}: Reinventing {RNN}s for the transformer era.
\newblock In H.~Bouamor, J.~Pino, and K.~Bali, editors, \emph{Findings of the Association for Computational Linguistics: EMNLP 2023}, pages 14048--14077, Singapore, Dec. 2023. Association for Computational Linguistics.
\newblock \doi{10.18653/v1/2023.findings-emnlp.936}.
\newblock URL \url{https://aclanthology.org/2023.findings-emnlp.936/}.

\bibitem[Radford and Narasimhan(2018)]{Radford2018ImprovingLU}
A.~Radford and K.~Narasimhan.
\newblock Improving language understanding by generative pre-training.
\newblock 2018.

\bibitem[Radford et~al.(2019)Radford, Wu, Child, Luan, Amodei, Sutskever, et~al.]{radford2019language}
A.~Radford, J.~Wu, R.~Child, D.~Luan, D.~Amodei, I.~Sutskever, et~al.
\newblock Language models are unsupervised multitask learners.
\newblock \emph{OpenAI blog}, 1\penalty0 (8):\penalty0 9, 2019.

\bibitem[Sun et~al.(2024)Sun, Li, Dalal, Xu, Vikram, Zhang, Dubois, Chen, Wang, Koyejo, Hashimoto, and Guestrin]{sun2024learninglearntesttime}
Y.~Sun, X.~Li, K.~Dalal, J.~Xu, A.~Vikram, G.~Zhang, Y.~Dubois, X.~Chen, X.~Wang, S.~Koyejo, T.~Hashimoto, and C.~Guestrin.
\newblock Learning to (learn at test time): Rnns with expressive hidden states, 2024.
\newblock URL \url{https://arxiv.org/abs/2407.04620}.

\bibitem[Villalobos et~al.(2024)Villalobos, Ho, Sevilla, Besiroglu, Heim, and Hobbhahn]{villalobos2024position}
P.~Villalobos, A.~Ho, J.~Sevilla, T.~Besiroglu, L.~Heim, and M.~Hobbhahn.
\newblock Position: Will we run out of data? limits of {LLM} scaling based on human-generated data.
\newblock In \emph{Forty-first International Conference on Machine Learning}, 2024.
\newblock URL \url{https://openreview.net/forum?id=ViZcgDQjyG}.

\bibitem[Wang et~al.(2024{\natexlab{a}})Wang, Yao, Xu, Qiao, Deng, Wang, Chen, Gu, Jiang, Xie, Huang, Chen, and Zhang]{wang-etal-2024-knowledge-mechanisms}
M.~Wang, Y.~Yao, Z.~Xu, S.~Qiao, S.~Deng, P.~Wang, X.~Chen, J.-C. Gu, Y.~Jiang, P.~Xie, F.~Huang, H.~Chen, and N.~Zhang.
\newblock Knowledge mechanisms in large language models: A survey and perspective.
\newblock In Y.~Al-Onaizan, M.~Bansal, and Y.-N. Chen, editors, \emph{Findings of the Association for Computational Linguistics: EMNLP 2024}, pages 7097--7135, Miami, Florida, USA, Nov. 2024{\natexlab{a}}. Association for Computational Linguistics.
\newblock \doi{10.18653/v1/2024.findings-emnlp.416}.
\newblock URL \url{https://aclanthology.org/2024.findings-emnlp.416/}.

\bibitem[Wang et~al.(2024{\natexlab{b}})Wang, Feng, Wang, Shi, Balachandran, He, and Tsvetkov]{wang2024resolving}
Y.~Wang, S.~Feng, H.~Wang, W.~Shi, V.~Balachandran, T.~He, and Y.~Tsvetkov.
\newblock Resolving knowledge conflicts in large language models.
\newblock In \emph{First Conference on Language Modeling}, 2024{\natexlab{b}}.
\newblock URL \url{https://openreview.net/forum?id=ptvV5HGTNN}.

\bibitem[Weidinger et~al.(2021)Weidinger, Mellor, Rauh, Griffin, Uesato, Huang, Cheng, Glaese, Balle, Kasirzadeh, Kenton, Brown, Hawkins, Stepleton, Biles, Birhane, Haas, Rimell, Hendricks, Isaac, Legassick, Irving, and Gabriel]{weidinger2021ethicalsocialrisksharm}
L.~Weidinger, J.~Mellor, M.~Rauh, C.~Griffin, J.~Uesato, P.-S. Huang, M.~Cheng, M.~Glaese, B.~Balle, A.~Kasirzadeh, Z.~Kenton, S.~Brown, W.~Hawkins, T.~Stepleton, C.~Biles, A.~Birhane, J.~Haas, L.~Rimell, L.~A. Hendricks, W.~Isaac, S.~Legassick, G.~Irving, and I.~Gabriel.
\newblock Ethical and social risks of harm from language models, 2021.
\newblock URL \url{https://arxiv.org/abs/2112.04359}.

\bibitem[Xu et~al.(2024)Xu, Qi, Guo, Wang, Wang, Zhang, and Xu]{xu-etal-2024-knowledge-conflicts}
R.~Xu, Z.~Qi, Z.~Guo, C.~Wang, H.~Wang, Y.~Zhang, and W.~Xu.
\newblock Knowledge conflicts for {LLM}s: A survey.
\newblock In Y.~Al-Onaizan, M.~Bansal, and Y.-N. Chen, editors, \emph{Proceedings of the 2024 Conference on Empirical Methods in Natural Language Processing}, pages 8541--8565, Miami, Florida, USA, Nov. 2024. Association for Computational Linguistics.
\newblock \doi{10.18653/v1/2024.emnlp-main.486}.
\newblock URL \url{https://aclanthology.org/2024.emnlp-main.486/}.

\bibitem[Yang et~al.(2024)Yang, Gribovskaya, Kassner, Geva, and Riedel]{yang-etal-2024-large-language-models}
S.~Yang, E.~Gribovskaya, N.~Kassner, M.~Geva, and S.~Riedel.
\newblock Do large language models latently perform multi-hop reasoning?
\newblock In L.-W. Ku, A.~Martins, and V.~Srikumar, editors, \emph{Proceedings of the 62nd Annual Meeting of the Association for Computational Linguistics (Volume 1: Long Papers)}, pages 10210--10229, Bangkok, Thailand, Aug. 2024. Association for Computational Linguistics.
\newblock \doi{10.18653/v1/2024.acl-long.550}.
\newblock URL \url{https://aclanthology.org/2024.acl-long.550/}.

\bibitem[Yao et~al.(2023)Yao, Wang, Tian, Cheng, Li, Deng, Chen, and Zhang]{yao-etal-2023-editing}
Y.~Yao, P.~Wang, B.~Tian, S.~Cheng, Z.~Li, S.~Deng, H.~Chen, and N.~Zhang.
\newblock Editing large language models: Problems, methods, and opportunities.
\newblock In H.~Bouamor, J.~Pino, and K.~Bali, editors, \emph{Proceedings of the 2023 Conference on Empirical Methods in Natural Language Processing}, pages 10222--10240, Singapore, Dec. 2023. Association for Computational Linguistics.
\newblock \doi{10.18653/v1/2023.emnlp-main.632}.
\newblock URL \url{https://aclanthology.org/2023.emnlp-main.632/}.

\bibitem[Yi et~al.(2024)Yi, Ye, Chen, Zhu, Chen, Lian, Sun, Xie, and Wu]{yi-etal-2024-vulnerability}
J.~Yi, R.~Ye, Q.~Chen, B.~Zhu, S.~Chen, D.~Lian, G.~Sun, X.~Xie, and F.~Wu.
\newblock On the vulnerability of safety alignment in open-access {LLM}s.
\newblock In L.-W. Ku, A.~Martins, and V.~Srikumar, editors, \emph{Findings of the Association for Computational Linguistics: ACL 2024}, pages 9236--9260, Bangkok, Thailand, Aug. 2024. Association for Computational Linguistics.
\newblock \doi{10.18653/v1/2024.findings-acl.549}.
\newblock URL \url{https://aclanthology.org/2024.findings-acl.549/}.

\bibitem[Zhang et~al.(2025)Zhang, Ye, Liu, Wu, Ren, and Chen]{zhang2025uncovering}
M.~Zhang, X.~Ye, Q.~Liu, S.~Wu, P.~Ren, and Z.~Chen.
\newblock Uncovering overfitting in large language model editing.
\newblock In \emph{The Thirteenth International Conference on Learning Representations}, 2025.
\newblock URL \url{https://openreview.net/forum?id=t8qcGXaepr}.

\bibitem[Zhang et~al.(2024)Zhang, Yao, Tian, Wang, Deng, Wang, Xi, Mao, Zhang, Ni, Cheng, Xu, Xu, Gu, Jiang, Xie, Huang, Liang, Zhang, Zhu, Zhou, and Chen]{zhang2024comprehensivestudyknowledgeediting}
N.~Zhang, Y.~Yao, B.~Tian, P.~Wang, S.~Deng, M.~Wang, Z.~Xi, S.~Mao, J.~Zhang, Y.~Ni, S.~Cheng, Z.~Xu, X.~Xu, J.-C. Gu, Y.~Jiang, P.~Xie, F.~Huang, L.~Liang, Z.~Zhang, X.~Zhu, J.~Zhou, and H.~Chen.
\newblock A comprehensive study of knowledge editing for large language models, 2024.
\newblock URL \url{https://arxiv.org/abs/2401.01286}.

\bibitem[Zhong et~al.(2023)Zhong, Wu, Manning, Potts, and Chen]{zhong-etal-2023-mquake}
Z.~Zhong, Z.~Wu, C.~Manning, C.~Potts, and D.~Chen.
\newblock {MQ}u{AKE}: Assessing knowledge editing in language models via multi-hop questions.
\newblock In H.~Bouamor, J.~Pino, and K.~Bali, editors, \emph{Proceedings of the 2023 Conference on Empirical Methods in Natural Language Processing}, pages 15686--15702, Singapore, Dec. 2023. Association for Computational Linguistics.
\newblock \doi{10.18653/v1/2023.emnlp-main.971}.
\newblock URL \url{https://aclanthology.org/2023.emnlp-main.971/}.

\bibitem[Zhu et~al.(2025)Zhu, Huang, Zhang, Jordan, Jiao, Tian, and Russell]{zhu2025towards}
H.~Zhu, B.~Huang, S.~Zhang, M.~Jordan, J.~Jiao, Y.~Tian, and S.~J. Russell.
\newblock Towards a theoretical understanding of the'reversal curse'via training dynamics.
\newblock \emph{Advances in Neural Information Processing Systems}, 37:\penalty0 90473--90513, 2025.

\end{thebibliography}

% \appendix

%%%%%%%%%%%%%%%%%%%%%%%%%%%%%%%%%%%%%%%%%%%%%%%%%%%%%%%%%%%%%%%%%%%%%%%%%%%%%%%
%%%%%%%%%%%%%%%%%%%%%%%%%%%%%%%%%%%%%%%%%%%%%%%%%%%%%%%%%%%%%%%%%%%%%%%%%%%%%%%
% % APPENDIX
% %%%%%%%%%%%%%%%%%%%%%%%%%%%%%%%%%%%%%%%%%%%%%%%%%%%%%%%%%%%%%%%%%%%%%%%%%%%%%%%
% %%%%%%%%%%%%%%%%%%%%%%%%%%%%%%%%%%%%%%%%%%%%%%%%%%%%%%%%%%%%%%%%%%%%%%%%%%%%%%%
% \newpage
% \appendix

% \section{Proof of theorems}

\end{document}